\documentclass[conference]{IEEEtran}
\IEEEoverridecommandlockouts

\usepackage{amsmath, bbm}
\usepackage{amssymb}
\usepackage{amsfonts}
\usepackage{amsthm}
\usepackage{algorithm}
\usepackage{algorithmic}
\usepackage{nicefrac}
\usepackage{bm}
\usepackage{hyperref}
\usepackage{cite}
\usepackage[utf8]{inputenc} 
\usepackage[T1]{fontenc}
\usepackage{graphicx}
\usepackage[font=small]{caption}
\usepackage{tikz}
\usetikzlibrary{positioning}
\usepackage{pgfplots}
\usepackage{epstopdf}
\usepackage[labelformat=simple]{subcaption}
\usepackage{multirow}
\usepackage{lipsum}
\usepackage{mathtools}

\NewDocumentCommand{\inR}{ m o }{\in \mathbb{R}^{#1 \IfValueT{#2}{\times #2}}}

\newcommand{\fracpart}[2]{\frac{\partial #1}{\partial #2}}

\theoremstyle{definition}

\makeatletter
\def\th@definition{
  \normalfont
  \itshape
}
\makeatother

\input{mysymbol.sty}

\begin{document}
\title{Fully Distributed Online Training of Graph Neural Networks in Networked Systems
\thanks{Research was sponsored by the DEVCOM ARL Army Research Office and was accomplished under Cooperative Agreement Number W911NF-19-2-0269. 
The views and conclusions contained in this document are those of the authors and should not be interpreted as representing the official policies, either expressed or implied, of the DEVCOM ARL Army Research Office or the U.S. Government. 
The U.S. Government is authorized to reproduce and distribute reprints for Government purposes notwithstanding any copyright notation herein.
\newline
Emails: $^\star$\{rostyslav.olshevskyi, zhongyuan.zhao, segarra\}@rice.edu, $^\ddag$\{kevin.s.chan.civ, gunjan.verma.civ, ananthram.swami.civ\}@army.mil}
\thanks{Source code and data: \url{https://github.com/RostyslavUA/fdTrainGNN}}
}

\author{
Rostyslav~Olshevskyi$^\star$, Zhongyuan~Zhao$^\star$, Kevin Chan$^\ddag$, Gunjan Verma$^\ddag$, Ananthram Swami$^\ddag$, Santiago~Segarra$^\star$\\
\textit{$^\star$Rice University, USA \hspace{10mm}  \hspace{2mm}  $^\ddag$DEVCOM Army Research Laboratory, USA}}
\maketitle

\begin{abstract}
Graph neural networks (GNNs) are powerful tools for developing scalable, decentralized artificial intelligence in large-scale networked systems, such as wireless networks, power grids, and transportation networks. 
Currently, GNNs in networked systems mostly follow a paradigm of `centralized training, distributed execution', which limits their adaptability and slows down their development cycles. 
In this work, we fill this gap for the first time by developing a communication-efficient, fully distributed online training approach for GNNs applied to large networked systems.
For a mini-batch with $B$ samples, our approach of training an $L$-layer GNN only adds $L$ rounds of message passing to the $LB$ rounds required by GNN inference, with doubled message sizes.
Through numerical experiments in graph-based node regression, power allocation, and link scheduling in wireless networks, we demonstrate the effectiveness of our approach in training GNNs under supervised, unsupervised, and reinforcement learning paradigms. 
\end{abstract}


%
\IEEEpeerreviewmaketitle

\section{Introduction}

Graph neural networks (GNNs) hold the promise of empowering networked artificial intelligence in communication networks, smart grids, and transportation networks, due to several unique features~\cite{wu2020comprehensive,chien2024opportunities}: 
1) \emph{permutation equivariance} as an important inductive bias for tasks in networks, 
2) \emph{local message passing (MP)} that naturally promotes distributed executions, and 
3) \emph{shared trained model among all nodes}, which allows GNNs to generalize and scale up to large, dynamic networks much easier than, e.g., multi-agent reinforcement learning (MARL)~\cite{nasir2019marl}. 
GNNs have been applied to enhance resource allocation and decision-making in wireless networks, such as power allocation, link scheduling, packet routing, network simulation and management, and computation offloading~\cite{chowdhury2021uwmmse,zhao2022twc,zhao2024bbp,zhao2024offloading,zhao2023graphbased,shen2023gnnforwireless,li2024glance}, by leveraging their ability to exploit the topological information of the connectivity and interference relationships among wireless devices. 

Current applications of GNNs in networked systems follow a paradigm of `\emph{centralized training, distributed execution}'. 
In particular, the centralized training of GNNs can only be done offline in simulated environments, requiring extensive efforts in data collection and environment modeling as well as computing resources.
Moreover, the distribution and deployment of trained models may cause downtime and disruptions to the networked system.
After deployment, the trained models would also likely experience distribution shifts due to mismatched training settings, changing real-world environments, and application scenarios. 
Therefore, fully distributed online training of GNNs could simplify the development of intelligent networked systems and improve their adaptivity.

Existing approaches in distributed machine learning \cite{verbraeken2020dml} are inadequate for fully distributed online training of GNNs. 
For example, with MARL~\cite{nasir2019marl}, each agent has a different trained model rather than a common trained model shared among all nodes as in GNNs. 
Although federated learning seeks to train a common model for many clients with the help of a central server~\cite{mcmahan2017communication}, neither the training nor inference requires any interactions between these clients, which is in contrast with GNNs where the inference requires synchronized message exchanging between each node and its neighbors. 
Distributed optimization (DO)~\cite{nedic2009distributed,swenson2022distributed,nazari2022dadam,chen2023dams} leverages many connected workers to accelerate the training of a model by splitting the training dataset and exchanging gradients among the workers. 
However, each worker in DO can perform model inference and backpropagation individually, which is different from GNNs where both inference and backpropagation require synchronized collaboration among all nodes in the network. 
{Existing works on distributed training of GNNs can be categorized as DO, where large graphs are divided into smaller subgraphs, which are distributed to different servers for memory and computing efficiencies~\cite{Lin2023survey,Shao2024distributed}.
By contrast, in our \emph{fully} distributed training, every node in the graph is its own computing server.
Thus, communication across servers is required for both inference and training.
}


{To fill the gap of training GNNs online in a fully distributed manner,
we take the supervised learning for graph convolutional neural networks (GCNNs)~\cite{gama2019gcnn,kipf2017semisupervised} as an example, transform it into a variation of the DO framework, and develop communication-efficient implementations based on local MP.
These principles of distributed training can serve as the basis for other types of GNNs~\cite{wu2020comprehensive}, such as edge-featured GNNs and graph attention networks (GATs), as well as unsupervised training and reinforcement learning for GNNs in sophisticated algorithmic frameworks~\cite{chowdhury2021uwmmse,zhao2023graphbased}.}

\noindent\textbf{Contributions:} Our contributions are as follows:
\begin{itemize}
\item We show that GNN training can be reformulated as a DO problem by decomposing the global objective, loss function, and gradient of GNNs into linear combinations of the corresponding local ones, and deriving a local form of backpropagation for GCNNs. 
\item We develop a communication-efficient approach for distributed training of GNNs, by incorporating distributed gradient descent schemes, rearranging gradient aggregation, and message piggybacking in mini-batch settings. 
\item Through numerical experiments, we demonstrate that our distributed training scheme not only achieves a convergence behavior very close to that of the classical, centrally-trained approach in supervised learning, but is also effective in more sophisticated ML pipelines such as graph-based algorithmic unfolding~\cite{chowdhury2021uwmmse} and graph-based actor-critic reinforcement learning frameworks~\cite{zhao2023graphbased}. 
\end{itemize}



\noindent\textbf{Notational convention:} 
$ (\cdot)^\top $, $ \odot $, and $ |\cdot| $ represent the transpose operator, Hadamard (element-wise) product operator, and the cardinality of a set {(or dimensionality of a vector)}, respectively.
$ \mathbb{E}(\cdot) $ stands for expectation.
Calligraphic symbols, e.g., $\ccalV$, denote a set. 
Upright bold lower-case symbols, e.g., $\bbz$, denote a column vector and $\bbz_i$ denotes the $i$-th element of vector $\bbz$. 
Upright bold upper-case symbols, e.g., $\bbZ$ denote a matrix, of which the element at row $i$ and column $j$ is denoted by $\bbZ_{ij}$, the row $i$ by $\bbZ_{i*}$, and the column $j$ by $\bbZ_{*j}$.


\section{Problem Formulation} \label{sec:system}

Consider a connected and undirected graph $\ccalG=(\ccalV, \ccalE)$, where $\ccalV$ is the set of nodes, $\ccalE$ is the set of edges, matrix $\bbX\in\reals^{|\ccalV|\times g_0}$ collects node features (e.g., the type or queueing state of a transmitter), and vector $\bby\in\reals^{|\ccalV|}$ collects node-wise labels (e.g., optimal power). 
An $L$-layer 
GCNN is a parameterized function $\hat{\bby} = \Psi_{\ccalG}(\bbX;\bbtheta)$ defined on $\ccalG$, where vector $\hat{\bby}\in\reals^{|\ccalV|}$ is the node-wise prediction, and  $\bbtheta$ collects all the trainable parameters.
Centralized training of the GCNN with supervised learning can be formulated as minimizing the expected loss over the distribution of node-featured graphs and the corresponding label vectors, $(\ccalG,\bbX,\bby)\in\Omega$, as
\begin{subequations}\label{E:central}
	\begin{align}
		\bbtheta^* &= \argmin_{\bbtheta\in\reals^{|\bbtheta|}} J(\bbtheta) \label{E:central:min}\\
		\text{s.t. } 
		  J(\bbtheta) & = \mathbb{E}_{(\ccalG,\bbX,\bby)\in\Omega}\left[\ell(\bby,\ccalG,\bbX;\bbtheta)\right]\;, \label{E:central:obj} \\
            \ell(\bbtheta) & =\ell(\bby, \ccalG, \bbX; \bbtheta) = \frac{1}{|\ccalV|}\sum_{i\in\ccalV}(\bby_i - \hat{\bby}_i)^2\;,\label{E:central:mse}\\
            \hat{\bby} &= \Psi_{\ccalG}(\bbX;\bbtheta),\label{E:central:gcn}
	\end{align} 
\end{subequations}
where \eqref{E:central:mse} defines the mean-squared-error (MSE) loss function.
Problem~\eqref{E:central} can be solved with stochastic gradient descent (SGD) with a learning rate of $\alpha$,
\begin{equation}\label{E:sgd}
    \bbtheta \!\leftarrow\! \bbtheta \!-\! \alpha \widehat{\nabla J(\bbtheta)},
    \;\widehat{\nabla J(\bbtheta)} \!=\! \nabla \ell(\bbtheta) \!=\! \left[\frac{\partial \ell( \bbtheta) }{\partial \bbtheta}\right]^{\top}\in\reals^{|\bbtheta|}\;.
\end{equation}
The local implementation of a GCNN comprises synchronized parallel executions of the same parameterized local function on every node, where each local function performs $L$ iterations of a local neighborhood aggregation followed by a dense layer.
Thus, the $L$-layer GCNN can be denoted as $ \hat{\bby} = \Psi_{\ccalG}(\bbX;\{\bbtheta^{i}\}_{i\in\ccalV}) $, where $\bbtheta^{i}=\bbtheta$ is a local copy of the global parameters on node $i\in\ccalV$.
The backward pass of a GCNN first computes the partial derivatives $ \partial \ell(\bbtheta)/\partial \bbtheta^i $ for all $i\in\ccalV$, and then the total derivative as 
\begin{equation}\label{E:agg}
    \frac{\partial \ell(\bbtheta)}{\partial \bbtheta} = \sum_{i\in\ccalV} \frac{\partial \ell(\bbtheta)}{\partial \bbtheta^{i}}\;.
\end{equation}

For the centralized training of a GCNN, the operation in \eqref{E:agg} is straightforward since the global loss function $\ell(\bbtheta)$ and all local gradients reside on the same server. 
{However, for GCNNs in fully distributed systems, 
where each node $i\in\ccalV$ is an individual device in the network,} online training becomes challenging as it requires a central server to host the global loss function in~\eqref{E:central:mse}, perform the summation operation in \eqref{E:agg}, SGD in~\eqref{E:sgd}, and redistribution of $\bbtheta$ to $\bbtheta^{i}$ for all $i\in\ccalV$.

To transform the centralized training in \eqref{E:central} into a distributed problem, we define the local loss $\ell_i(\bbtheta)$ and local objective $J_i(\bbtheta)$ for all $i\in\ccalV$ as follows, 
\begin{equation}\label{E:localloss}
    \ell(\bbtheta) = \frac{1}{|\ccalV|}\sum_{i\in\ccalV} \ell_i(\bbtheta)\;, \text{where}\; \ell_i(\bbtheta)=({\bby_i-\hat{\bby}_i})^2\;,
\end{equation}
\begin{equation}\label{E:localobj}
    J(\bbtheta) = \frac{1}{|\ccalV|}\sum_{i\in\ccalV}J_i(\bbtheta)\;, \text{where}\; J_i(\bbtheta) = \mathbb{E}_{\Omega}\left[\ell_i(\bbtheta)\right]\;.
\end{equation}
With \eqref{E:localloss}, \eqref{E:localobj}, and a small $\epsilon>0$, we reformulate \eqref{E:central} for {a sampling distribution $\Omega_{\ccalV}$ conditioned on a  vertex set $\ccalV$} as
\begin{subequations}\label{E:dist}
	\begin{align}
		\{\bbtheta^i\}_{i\in\ccalV}^* &= \argmin_{\bbtheta^i\in\reals^{|\bbtheta|}, i\in\ccalV} \frac{1}{|\ccalV|}\sum_{i\in\ccalV}J_i(\bbtheta) \label{E:dist:obj}\\
		\text{s.t. } 
		  J_i(\bbtheta) &= \mathbb{E}_{\Omega_{\ccalV}}\left[\ell_i(\bbtheta)\right]\;, \label{E:dist:cost} \\
    \ell_i(\bbtheta)&=({\bby_i-\hat{\bby}_i})^2\;,\forall\; i\in\ccalV\;,\label{E:dist:se}\\
    \hat{\bby} &= \Psi_{\ccalG}(\bbX;\{\bbtheta^{i}\}_{i\in\ccalV}),\; (\ccalG,\bbX,\bby)\in\Omega_{\ccalV}\;,\label{E:dist:gcn}\\
        \bbtheta &=\frac{1}{|\ccalV|}\sum_{i\in\ccalV}\bbtheta^i,\;  \lVert{\bbtheta^i - \bbtheta}\rVert^2_2 < \epsilon,\; \forall\; i\in\ccalV .\label{E:dist:equal}
	\end{align} 
\end{subequations}
Based on \eqref{E:agg} and \eqref{E:localloss}, the total derivative becomes 
\begin{equation}\label{E:localgrad}
    \frac{\partial \ell(\bbtheta)}{\partial \bbtheta} = \frac{1}{|\ccalV|}\sum_{i\in\ccalV}\sum_{j\in\ccalV} \frac{\partial \ell_j(\bbtheta)}{\partial \bbtheta^{i}}\;.
\end{equation}
Consequently, the SGD in \eqref{E:sgd} can be implemented in a distributed manner as
\begin{subequations}\label{E:grad}
	\begin{align}
 \bbtheta^i &\leftarrow\bbtheta^i-\alpha\widehat{\nabla J(\bbtheta)}\;,\forall\; i\in\ccalV\;,\label{E:grad:sgd}\\
\widehat{\nabla J(\bbtheta)} &= \frac{1}{|\ccalV|}\sum_{i\in\ccalV}\widehat{\nabla J_i(\bbtheta)}\;,\label{E:grad:avg}\\
\widehat{\nabla J_i(\bbtheta)} &= \left[\sum_{j\in\ccalV} \frac{\partial \ell_j(\bbtheta)}{\partial \bbtheta^{i}}\right]^{\top},\forall\; i\in\ccalV\;,\label{E:grad:local}
	\end{align} 
\end{subequations}
given that $\bbtheta^i=\bbtheta$, for all $i\in\ccalV$ upon initialization.

{Unlike the centralized training in \eqref{E:central}, the distributed optimization in~\eqref{E:dist} no longer requires a server to host the global loss function $\ell(\bbtheta)$ or global objective function $J(\bbtheta)$.}
Moreover, as long as \eqref{E:grad:avg} and \eqref{E:grad:local} can be computed in a fully distributed manner, the centralized SGD in~\eqref{E:sgd} and~\eqref{E:agg} can be attained via \eqref{E:grad} without the need of a central server.


\vspace{1mm}
\noindent
{\bf Key departure from classical DO.}
It is essential to notice that in a local implementation of an $L$-layer GCNN 
where every node $i\in\ccalV$ has a copy of the parameters $\bbtheta^i$, the estimate $\hat{\bby}_i$ at node $i$ is  \emph{not only a function of} $(\bbX_{i*},\bbtheta^i)$ but also those of its $L$-hop neighbors,  
$$ \hat{\bby}_i = f_{\text{local}}\left( \left\{\bbX_{j*};\bbtheta^j\right\}_{j\in\ccalN^{L}_{\ccalG}(i)\cup \{i\}}\right) \;,$$
where $ \ccalN^{L}_{\ccalG}(i) $ denotes the set of $L$-hop neighbors of node $i$ on graph $\ccalG$. 
This follows immediately from the fact that $\hat{\bby}_i$ depends on messages that node $i$ receives from its neighbors $j$, and these messages are functions of their parameters $\{\bbtheta^j\}$. 
This dependence cascades over the $L$ layers, as later illustrated in \eqref{E:gcn:local} for the specific case of GCNNs. 
Consequently, unlike in classical DO~\cite{nedic2009distributed,swenson2022distributed,nazari2022dadam,chen2023dams}, node $i$ \emph{cannot immediately compute the local gradient} since $\frac{\partial \ell(\bbtheta)}{\partial \bbtheta^i} \neq \frac{1}{|\ccalV|} \frac{\partial \ell_i(\bbtheta)}{\partial \bbtheta^{i}}$.
In this context, the local gradient computation requires further consideration.
In the next section, we introduce our fully distributed solution to implement~\eqref{E:grad}.
Our objective is twofold: first, we want to minimize the global objective function in~\eqref{E:dist:obj} in a fully distributed manner.
Second, we aim to minimize the communication costs of our fully distributed training. 


\section{Fully Distributed Training of GNNs}\label{sec:solution}

Our solution comprises three components: 1)~fully-distributed backpropagation for local gradient estimation~\eqref{E:grad:local}  in Section~\ref{sec:solution:bp};
2) joint implementation of \eqref{E:grad:avg} and \eqref{E:grad:sgd} based on distributed gradient descent approaches in Section \ref{sec:solution:do};
and 3) systematic schemes to reduce the communication rounds for mini-batch training in Section {\ref{sec:sol:eff_mbt}}. 

\subsection{{Fully Distributed Backpropagation}}\label{sec:solution:bp}

The $l$th layer of the GCNN $\hat{\bby} = \Psi_{\ccalG}(\bbX;\bbtheta)$ introduced in Section~\ref{sec:system}, where $l\in\{1,\dots,L\}$, can be expressed as
\begin{equation}\label{E:gcn}
	\mathbf{X}^{l} = \sigma_l\left(\mathbf{X}^{l-1}{\bbTheta}_{0}^{l}+\bbS \mathbf{X}^{l-1}{\bbTheta}_{1}^{l}\right), \; l\in\{1,\dots,L\},
\end{equation}
where $\bbX^l \in\reals^{|\ccalV|\times g_{l}}$ collects the output node features of layer $l$, matrices $ \bbTheta_{0}^{l}, \bbTheta_{1}^{l}\in\reals^{g_{l-1}\times g_{l}} $ are the trainable parameters of the $l$th layer, $\sigma_{l}$ is an element-wise activation function, and $\bbS\in\reals^{|\ccalV|\times |\ccalV|}$ is the graph shift operator.
$\bbS$ can be selected as the adjacency matrix $\bbA$, the graph Laplacian $\bbL$, or their normalized versions~\cite{gama2019gcnn}.
For the GCNN, the input node feature $\bbX^0 = \bbX$, and prediction $\hat{\bby} = \bbX^L $ where $g_L=1$.

The local form of the GCNN layer in \eqref{E:gcn} on node $i\in\ccalV$ is
\begin{equation}\label{E:gcn:local}
    \bbX_{i*}^{l} \!=\! \sigma_l (\bbH_{i*}^{l})\;,\;  \bbH_{i*}^{l} \!= \!\bbX_{i*}^{l-1} \bbTheta_{0}^{li} +\!\!\!\!\sum_{j \in \mathcal{N}^{+}_{\ccalG}(i)}\!\!\!\!{\bbS_{ij}}{\bbX_{j*}^{l-1}} \bbTheta_{1}^{li}\;,
\end{equation}
where $\bbX_{i*}^{l}\in\reals^{1\times g_{l}}$ captures the output features of layer $l$ on node $i$,
$\bbTheta_{0}^{li}$ and $\bbTheta_{1}^{li}$ are local copies at node $i$ of $\bbTheta_{0}^{l}$ and $\bbTheta_{1}^{l}$, and
$\mathcal{N}^{+}_{\ccalG}(i)=\ccalN_{\ccalG}(i)\cup\{i\}$ with $\ccalN_{\ccalG}(i) = \ccalN^1_{\ccalG}(i)$.
The distributed execution of a GCNN layer can be implemented in two steps: first, node $i\in\ccalV$ exchanges its input node feature $\bbX_{i*}^{l-1}$ with its neighbors $j\in\ccalN_{\ccalG}(i)$; second, each node 
locally computes \eqref{E:gcn:local}.
The $L$-layer GCNN requires $L$ rounds of MP.

To find the messages passed in backpropagation,
we define $\bbZ^{l}={\partial\ell(\bbtheta)}/{\partial\bbX^{l}}$ and $\bbQ^{l}={\partial\ell(\bbtheta)}/{\partial\bbH^{l}}$ for $l\in\{1,\dots,L\}$, where $\bbZ^{l},\bbQ^{l}\in\reals^{g_l\times|\ccalV|}$. 
For node $i\in\ccalV $, we have
\begin{equation}\label{E:z}
\bbZ^l_{*i}\!=\!\fracpart{\ell(\bbtheta)}{\bbX^l_{i*}}\in\reals^{g_l\times1},\; \bbZ^{L}_{*i}\!=\!\frac{\partial \ell(\bbtheta)}{\partial {\hat{\bby}_{i}}} \!=\! 2({\hat{\bby}_i\!-\!\bby_i})\;,
\end{equation}
\begin{equation}\label{E:q}
    \bbQ^l_{*i}=\fracpart{\ell(\bbtheta)}{\bbX^l_{i*}}\fracpart{\bbX^l_{i*}}{\bbH^l_{i*}}=\bbZ^l_{*i}\odot\sigma_{l}'(\bbH^l_{i*})\in\reals^{g_l\times1},
\end{equation}
where $\sigma_{l}'(\cdot)$ is the element-wise derivative function of the activation $\sigma_{l}(\cdot)$.
Based on~\eqref{E:gcn:local}, the local partial derivatives for the trainable parameters at node $i\in\ccalV$ are:
\begin{subequations}\label{E:bp:theta}
    \begin{align}
        \fracpart{\ell(\bbtheta)}{\bbTheta^{li}_{0}} &= \fracpart{\ell(\bbtheta)}{\bbH^l_{i*}}\fracpart{\bbH^l_{i*}}{\bbTheta^{li}_{0}} = \bbQ^l_{*i}\bbX^{l-1}_{i*} \in\reals^{g_l\times g_{l-1}} \;,\label{E:bp:theta:0}\\
        \fracpart{\ell(\bbtheta)}{\bbTheta^{li}_{1}} &=\fracpart{\ell(\bbtheta)}{\bbH^l_{i*}}\fracpart{\bbH^l_{i*}}{\bbTheta^{li}_{1}} = \bbQ^l_{*i}\!\!\!\!\!\sum_{j \in \mathcal{N}^{+}_{\ccalG}(i)}\!\!\!\!{\bbS_{ij}}{\bbX_{j*}^{l-1}}  \;.\label{E:bp:theta:1}
    \end{align}
\end{subequations}
According to the chain rule, we can find $\bbZ^{l-1}_{*i}\in\reals^{g_{l-1}\times 1}$ as
\begin{equation}
\begin{aligned}
    \bbZ^{l-1}_{*i} &= \fracpart{\ell(\bbtheta)}{\bbX^{l-1}_{i*}} = \sum_{j\in\ccalV}\fracpart{\ell(\bbtheta)}{\bbH^{l}_{j*}}\fracpart{\bbH^{l}_{j*}}{\bbX^{l-1}_{i*}}\\
    &= \left(\bbTheta^{li}_{0}+{\bbS_{ii}} \bbTheta^{li}_{1}\right)\!\bbQ^l_{*i} + \!\!\!\! \sum_{j\in\ccalN_{\ccalG}(i)}\!\!\!{\bbS_{ji}}\bbTheta^{lj}_{1}\bbQ^l_{*j}\!\!,\label{E:bp:z}
\end{aligned}
\end{equation}
since, based on \eqref{E:gcn:local}, we have the following
\begin{equation*}
\left[\fracpart{\bbH^{l}_{j*}}{\bbX^{l-1}_{i*}}\right]^{\top} =
    \begin{cases}
        \bbTheta^{li}_{0}+{\bbS_{ii}} \bbTheta^{li}_{1} &\text{if } j=i \;,\\
        {\bbS_{ji}}\bbTheta^{lj}_{1} &\text{if } j\in\ccalN_{\ccalG}(i)\;, \\
        \mathbf{0}_{g_{l-1}\times g_l} & \text{if } j\notin\ccalN^{+}_{\ccalG}(i).\; 
    \end{cases}\;
\end{equation*}

Equations \eqref{E:z}-\eqref{E:bp:z} show the local form of backpropagation for all layers $l\in\{1,\dots,L\}$.
Notice that the sum operation in~\eqref{E:bp:theta:1} is already done in the forward pass in \eqref{E:gcn:local}. 
Only the second term in \eqref{E:bp:z} requires an additional round of MP, i.e., each node $i$ broadcasts $ \bbTheta^{li}_{1}\bbQ^l_{*i} $ to all its neighbors as ${\bbS_{ji}}$ can be found from the forward pass. 
Based on \eqref{E:z}-\eqref{E:bp:z}, $L-1$ rounds of MP are required to estimate the local gradient $\widehat{\nabla J_i(\bbtheta)} $ in \eqref{E:grad:local}.

\subsection{Distributed Stochastic Gradient Descent}\label{sec:solution:do}
A naive approach for implementing \eqref{E:grad:avg} is to perform $K\geq1$ rounds of distributed consensus on the local gradient estimates $\{\widehat{\nabla J_i(\bbtheta)}\}_{i\in\ccalV}$.
The $k$th round of distributed consensus on a set of node-specific 
vectors $\{\bbx^{j}(k)\}_{j\in\ccalV}$ can be expressed as
\begin{equation}\label{E:naive}
    \bbx^{i}(k+1) = \sum_{j\in\ccalN^{+}_{\ccalG}(i)}\!\!\! {\bbW_{ij}}\bbx^{j}(k)\;,\; k\in\{1,\dots, K\},
\end{equation}
where matrix $\bbW\in\reals^{|\ccalV|\times |\ccalV|}$ collects the consensus weights. 
Denoting the degree of node $i$ by $d(i)$, a good candidate for $\bbW$ is the Metropolis-Hasting weights~\cite{xiao2006distributed}
\begin{equation}\label{E:W}
    \bbW_{ij} = 
    \begin{cases}
    \frac{1}{1+\max\{d(i), d(j)\}} \quad &\{i,j\}\in\ccalE \;,\\
    1-\sum_{v\in\ccalN_{\ccalG}(i)}\bbW_{iv} & i=j \;,\\
    0 & \text{otherwise.}
    \end{cases}\;
\end{equation}
However, this naive approach has a high communication cost since a large $K$ is required for the convergence of the consensus operation in each gradient step.

Notice that the formulation in \eqref{E:dist:obj} can be considered as a form of DO but where the local gradient estimate $\widehat{\nabla J_i(\bbtheta)}$ is obtained as described in Section~\ref{sec:solution:bp}.
Thus, we can employ efficient approaches such as D-SGD~\cite{swenson2022distributed}, D-Adam~\cite{nazari2022dadam}, and D-AMSGrad~\cite{chen2023dams}. 
{In the context of the popular mini-batch gradient descent, where the network topology is assumed to be static during a mini-batch,}
each local update in DO comprises one round of distributed consensus on $\bbtheta^i$ and an application of local gradient,
\begin{equation}\label{E:dsgd}
    \bbtheta^i(t+1) =\!\!\!\! \sum_{j\in\ccalN^{+}_{\ccalG}(i)}\!\!\!\! {\bbW_{ij}}\bbtheta^{j}(t) - \alpha_t f\left(\left\{\widehat{\nabla J_i(\bbtheta)}(b)\right\}_{b=1}^{B} \right) \;,
\end{equation}
where the function $ f(\cdot) $ aggregates local gradients over a mini-batch of $B$ samples. 
{To estimate the gradients of $B$ samples, $\{\widehat{\nabla J(\bbtheta)}(b)\}_{b=1}^{B}$, we require $B$ forward passes of the GCNN for the inference of $ \{\hat{\bby}(b)\}_{b=1}^{B}$ and $B$ passes of backpropagation.}
For D-SGD, function $f(\cdot)$ is simply a summation $$f\left(\left\{\widehat{\nabla J_i(\bbtheta)}(b)\right\}_{b=1}^{B} \right)=\sum_{b=1}^{B} \widehat{\nabla J_i(\bbtheta)}(b)\;. $$
However, for D-Adam and D-AMSGrad, $ f(\cdot) $ is based on the momentum of the local gradients from graph data samples. 

The entire procedure of DO-based distributed training on {a network  of a fixed set of vertices $\ccalV$ with dynamic topology, i.e., edge set $\ccalE(t)$, is illustrated in Algorithm~\ref{alg:minibatch}, where $\Omega$ is the sampling distribution for $(\ccalG, \bbX, \bby)$, and $\Omega_{\ccalV}$ (or $\Omega_{\ccalG}$) is the conditional distribution for a given $\ccalV$ (or $\ccalG$).}

\begin{algorithm}[!t]
\caption{Distributed Training of GCNNs}\label{alg:minibatch}
\label{algo:foo}
\hspace*{\algorithmicindent} \textbf{Input}: ${\ccalV, \Omega,} \alpha, B$ \\
\hspace*{\algorithmicindent} \textbf{Output}: $\{\bbtheta^{i}\}^{*}_{i\in\ccalV}, {\bbtheta^*}$ 
\begin{algorithmic}[1] 
\STATE Initialize $\bbtheta$ randomly, $t\!=\!0$, $\{\bbtheta^{i}\!(t)\}_{i\in\ccalV}\!=\!\{\bbtheta\}_{i\in\ccalV}$, $\alpha_t\!=\!\alpha$ 
\WHILE{ not converged }
\STATE {Draw graph $\ccalG(t)\!=\!(\ccalV,\ccalE(t))$ {from $\Omega$ conditioned on ${\ccalV}$}}
\STATE {Compute $\bbW$ using \eqref{E:W}}
\FOR {$b=1,\dots, B$} 
\STATE Draw {node features and labels} $(\bbX(b),\bby(b))\!\sim\!\Omega_{\ccalG(t)}$
\FORALL{ $i\in\ccalV$}
\STATE Compute ${\hat{\bby}_i}(b)$ using~\eqref{E:gcn:local} for all $l\in\{1,\dots,L\}$.
\STATE Compute $\widehat{\nabla J_i(\bbtheta)} (b) $ using \eqref{E:z}--\eqref{E:bp:z}
\ENDFOR
\ENDFOR
\STATE Update {$\{\bbtheta^i(t+1)\}_{i\in\ccalV}$} using \eqref{E:dsgd}
\STATE $t=t+1$, update $\alpha_t$, e.g., via exponential decay
\ENDWHILE
\STATE {Return $ \{\bbtheta^{i}\}^{*}_{i\in\ccalV} = \{\bbtheta^{i}(t)\}_{i\in\ccalV} $, 
{$\bbtheta^*=\frac{1}{|\ccalV|}\sum_{i\in\ccalV}\bbtheta^i(t)$}}
\end{algorithmic}
\end{algorithm}




\subsection{Communication-efficient Mini-Batch Training}\label{sec:sol:eff_mbt}


If we approach~\eqref{E:grad:avg} naively by running distributed consensus on local gradients for each sample $(\bbX(b),\bby(b))$, 
the whole mini-batch requires $B(2L-1+K)$ rounds of MP.
Moreover, messages exchanged in distributed consensus are of large size, e.g., $|\bbtheta|$. 
We can reduce the communication cost by i) re-using information from the forward pass, ii) rearranging the mini-batch update, and iii) piggybacking messages for backward and forward passes, as we discuss next.

\begin{figure}[!t]
    \centering
    \vspace{-0.05in}
    \includegraphics[width=0.98\linewidth]{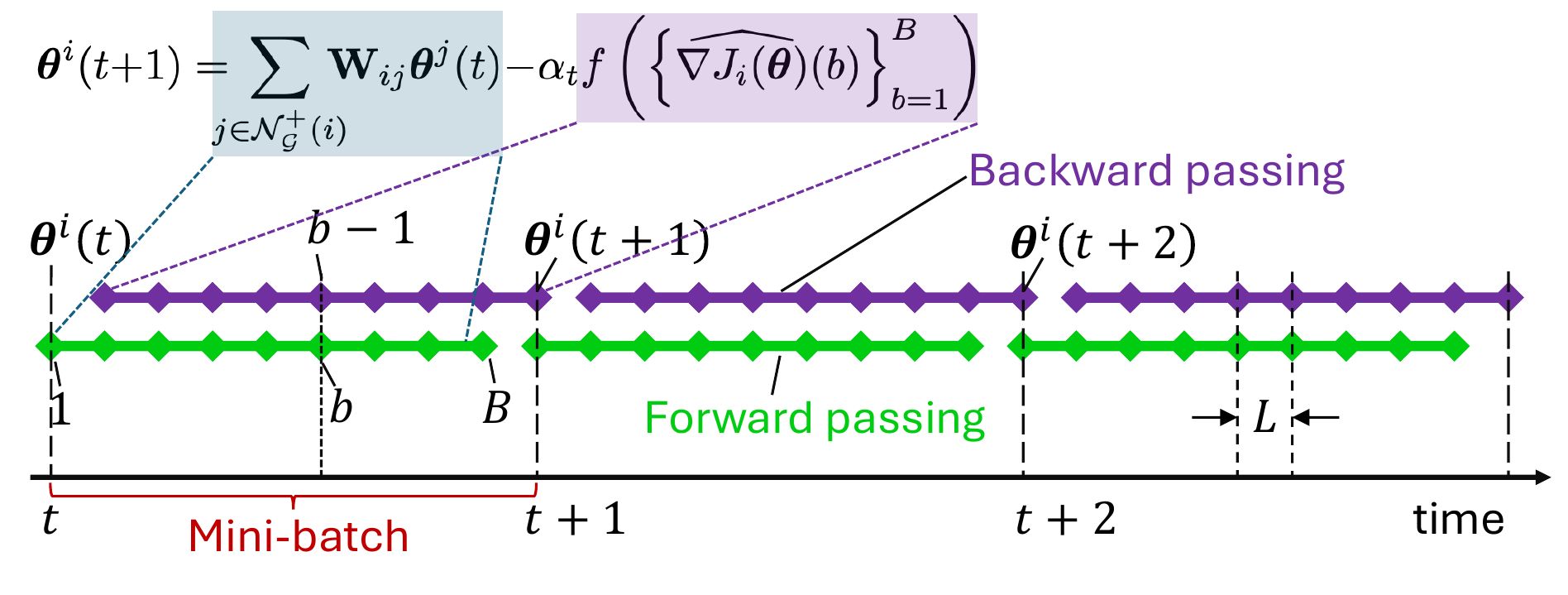}
    \vspace{-0.1in}
    \caption{Timeline of fully-distributed training of GCNN in mini-batches.  
    By piggybacking messages in the backward pass of sample $b-1$ into the messages of the forward pass of sample $b$, a mini-batch requires only $L(B+1)$ rounds of MP.
    {Notice that most communication and computation for the consensus step and local gradient aggregation (line $12$ in Algo~\ref{algo:foo}) can be piggybacked to the messages of $B$ forward passes and carried out in parallel with the processing of data samples (lines $5 - 11$ in Algo~\ref{algo:foo}). 
    }
    }
    \label{fig:dsgd_timeline}
    \vspace{-0.2in}
\end{figure}

\textit{Information reuse:} In the GCNN forward pass for sample $b$, node $i\in\ccalV$ can save the intermediate variables $\bbH^{l}_{i*}(b)$, $\bbX^{l}_{i*}(b)$ and $ \sum_{j \in \mathcal{N}^{+}_{\ccalG}(i)}{\bbS_{ij}}{\bbX_{j*}^{l-1}(b)} $ in \eqref{E:gcn:local} for all $l\in\{1,\dots,L\}$, which can be reused in \eqref{E:q} and \eqref{E:bp:theta} for the following backward pass, without retransmission and recomputing.

\textit{Mini-batch rearrangement:} Instead of running expensive distributed consensus for every sample, we can first aggregate $B$ local gradients at each node and then run $K$ rounds of distributed consensus once per mini-batch.
{The latter is mathematically equivalent to the former under basic SGD, 
$$
\sum_{b=1}^{B} \left[\frac{1}{|\ccalV|}\sum_{i\in\ccalV}\widehat{\nabla J_i(\bbtheta)}(b)\right] = \frac{1}{|\ccalV|}\sum_{i\in\ccalV}\left[\sum_{b=1}^{B}\widehat{\nabla J_i(\bbtheta)}(b)\right]\;,$$
but cuts the total rounds per mini-batch to $B(2L-1)+K$.
Furthermore, the DO-based gradient descent in~\eqref{E:dsgd} requires only $1$ round of consensus per mini-batch, where each node $i$ only needs to broadcast $\bbtheta^i(t)\in\reals^{|\bbtheta|}$ to its neighbors $j\in\ccalN_{\ccalG}(i) $ once.
Since local aggregation $f(\cdot)$ needs no MP, the required communication rounds is further reduced to $B(2L-1)+1$ per mini-batch.
In addition, for momentum-based DO, consensus on momentum parameters is also needed to ensure convergence, as in D-AMSGrad~\cite{chen2023dams}.} 

\begin{figure*}[!tbp]
  \hspace{-4mm}
  \subfloat[]{
  \includegraphics[height=1.62in]{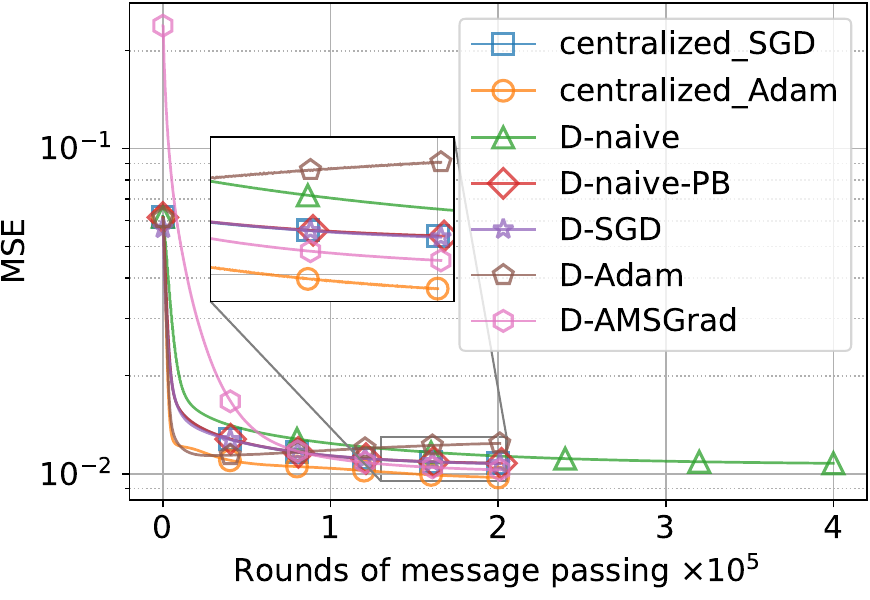}
  \label{fig:regression}\vspace{-0.1in}
  }
  \hspace{-4mm}
  \subfloat[]{
  \includegraphics[height=1.62in]{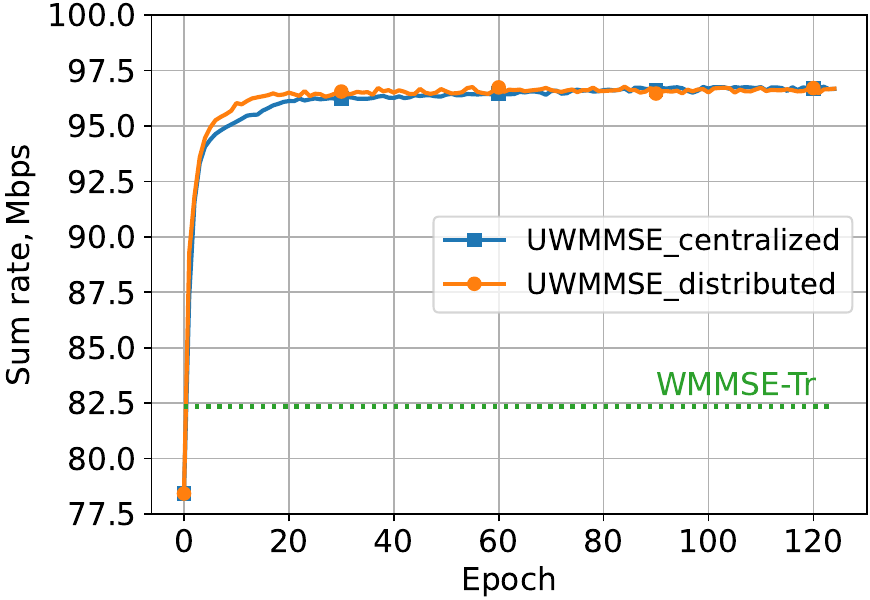}
  \label{fig:powalloc}\vspace{-0.1in}
  }
  \hspace{-4mm}
  \subfloat[]{
  \includegraphics[height=1.62in]{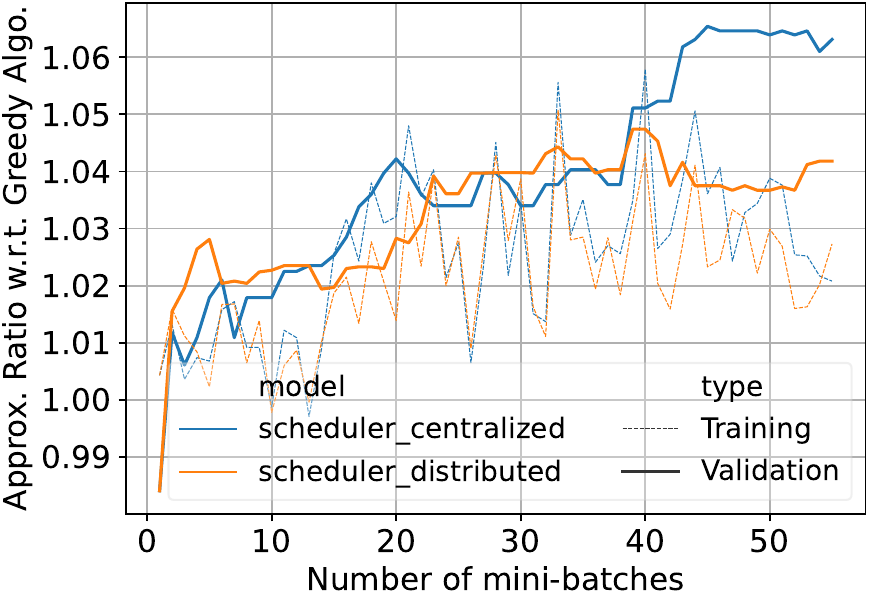}
  \label{fig:scheduling}\vspace{-0.1in}
  }
  \vspace{-0.1in}
  \caption{The evolution of objective values over the course of training: (a) Node regression, where a marker is placed every 200 mini-batches. 
  (b) Power allocation for a network of 25 transmitter-receiver pairs. 
  (c) Distributed link scheduling in conflict graphs of 100 nodes (links). 
  }
  \vspace{-0.1in}
\end{figure*}

\textit{Piggybacking:} 
Since it is more efficient to transmit a larger message than multiple smaller messages in wireless networks due to the signaling overhead of each transmission, we can piggyback the message of the backward pass for sample $b-1$ to the forward pass of the next sample $b$, as shown in Fig.~\ref{fig:dsgd_timeline}.
For example, in the $l$th round ($l<L$) of MP in the forward pass of sample $b$, each node $i\in\ccalV$ sends a message containing $ \bbX^{l}_{i*}(b) \in\reals^{1\times g_{l} } $ and $ [\bbTheta^{\bar{l}i}_{1} \bbQ^{\bar{l}i}_{*i}(b\!-\!1)]^{\top} \in \reals^{1\times g_{\bar{l}} }$, where $\bar{l}=L-l-1 $, to all its neighbors $j\in\ccalN_{\ccalG}(i)$, which can be achieved by a single broadcast transmission with an omnidirectional antenna.
When $l=L$ or $b=1$, only $\bbX^{l}_{i*}(b)$ is exchanged. 
Once per mini-batch, each node $i$ also needs to send $d(i)\in\reals$ and $\bbtheta^i(t)$ to all its neighbors $j\in\ccalN_{\ccalG}(i) $, which can be broken up and piggybacked in the messages of $b=1$ and $l=L, b>1$.
As a result, the total rounds of MP for a mini-batch are reduced to $LB+L-1$, where  $L-1$ rounds are for the backward pass of the last sample $b=B$.
The detailed communication costs for these approaches are listed in Table~\ref{tab:cost}.

\begin{table}[!t]
\vspace{-0.05in}
\caption{Communication cost of a mini-batch in GCNN training}\label{tab:cost}
\vspace{-0.05in}
\centering
\begin{tabular}{|p{3.2cm}|p{2.4cm}|p{1.8cm}|}
\hline
\textbf{Accumulated Measures} & \textbf{Total Msg. Rounds} & \textbf{Message size} \\  
\hline
Forward pass (FWD) only & $ LB $ & $ {g_l} $ \\ \hline
FWD + D-naive (grad. consensus per sample) + reuse & $ B(2L-1)$\newline $+BK $ & $ g_l $,\newline $|\bbtheta|$ \\ \hline
FWD + Grad. consensus per batch + Info. reuse & $ 2BL-B $\newline $+K$ & $ g_l $, \newline$|\bbtheta|$  \\ \hline
FWD + Grad. consensus per batch + reuse + piggyback & $ LB +L -1 $\newline $+K$ & $ g_l+g_{L-l-1} $,\newline $|\bbtheta|$  \\ \hline
FWD + Dist. Opt. in \eqref{E:dsgd} + reuse + piggyback & $ LB +L -1 $ & $ g_l+g_{L-l-1}+ $\newline $|\bbtheta|/LB$ \\
\hline
\end{tabular}
\end{table}

\section{Numerical Simulations}\label{sec:num_sim}
To evaluate the effectiveness of our proposed distributed training scheme, we compare it with centralized training (and naive distributed training) under supervised learning for node regression on a synthetic dataset (Section~\ref{ssec:synthetic_data}), unsupervised learning in UWMMSE wireless power allocation~\cite{chowdhury2021uwmmse} (Section~\ref{ssec:power_allocation}), and 
graph-based actor-critic reinforcement learning for distributed link scheduling~\cite{zhao2023graphbased} (Section~\ref{ssec:link_scheduling}).

\subsection{Supervised Learning on Synthetic Graph Data}
\label{ssec:synthetic_data}

In this experiment, we train a 2-layer GCNN to predict a set of graph data $ \{(\bbX(n),\bby(n))\in\Omega_{\ccalG}\}_{n=1}^{N}$ under a given random graph $\ccalG$ drawn from a Barab\'asi-Albert model ($m=2$) with $|\ccalV|=100$ nodes.
The input $\bbX\in\reals^{|\ccalV|\times 10}$ comprises a combination of continuous, binary, and discrete (one-hot) features drawn from uniform and normal distributions.
The labels are generated by a non-linear process as $\bby = \Phi_{\ccalG}(\bbX; \bbomega)+\bbn$, where $\Phi_{\ccalG}(\cdot)$ is a 2-layer GCNN parameterized by a set of random weights $\bbomega$, and $\bbn$ is Gaussian noise with a variance of $0.01$.
The hidden dimensions of the GCNN $\Phi$ are purposely chosen to be different from the GCNN $\Psi$ under training.
With a set of training samples $N= 1000$, batch size of $B=100$, a learning rate $\alpha=10^{-3}$, and a total of $1000$ epochs, 
we train a GCNN $\Psi_{\ccalG}(\cdot;\bbtheta)$ with supervised learning using different approaches.

The MSE loss as a function of MP rounds for 1000 epochs under 7 different training approaches are presented in Fig.~\ref{fig:regression}, where the differences in message sizes are ignored.
For centralized training (with SGD and Adam optimizers), only the cost of forward passing is considered such that processing each data sample requires $L=2$ rounds of communication.
The number of consensus rounds was set to $K=1$ for naive distributed approaches (D-naive and D-naive-PB, where PB stands for piggybacking).
After 1000 epochs, the MSE losses under all methods approach the noise floor of $0.01$. 
We consider centralized training with the SGD optimizer as the baseline.
The centralized Adam converges to the lowest MSE due to its momentum-based strategy.
D-SGD~\cite{swenson2022distributed} and D-naive-PB converge almost as well as the baseline, whereas D-naive takes twice as many rounds to achieve the same MSE.
D-Adam~\cite{nazari2022dadam} shows signs of overfitting after quick initial convergence due to the divergence of local momentum across nodes.
Among the distributed approaches, D-AMSGrad~\cite{chen2023dams} achieves the best convergence by incorporating distributed consensus on the local momentum terms.
Although D-naive-PB with $K=1$ requires a similar total number of rounds of MP as D-AMSGrad, there is one round of MP with large messages of size $|\bbtheta|$ at the end of each mini-batch, which may be translated to more rounds of exchanging normal-sized messages.
This result shows that D-AMSGrad is a good candidate for distributed training of GCNNs.


\subsection{Unsupervised Learning for UWMMSE Power Allocation}
\label{ssec:power_allocation}

A 4-layer neural architecture (UWMMSE) is constructed by unfolding 4 iterations of the weighted minimum mean-squared error (WMMSE) algorithm for power allocation, in which two constants are parameterized by two 2-layer GCNNs~\cite{chowdhury2021uwmmse}. 
Trained with unsupervised learning, the UWMMSE seeks to maximize the total throughput (sum rate) of 25 transmitter-receiver pairs by computing their transmit powers based on channel state information (CSI).
These 50 transceivers are randomly scattered in a 2D square with a width of $1\text{km}$, and operating on a $5$MHz band at the center frequency of $900$MHz, with a maximum power of $5$W.
The training dataset contains 640,000 realizations of CSI matrices under 100 realizations of transceiver locations, generated from the urban macro path loss model~\cite{3gppTR38901} with Rayleigh fading, representing a wireless network with node mobility.
We evaluate centralized (Adam) and distributed (D-Adam) training 
with a batch size of $B=64$, and a learning rate of $\alpha=10^{-2}$.
As shown in Fig. \ref{fig:powalloc}, UWMMSE outperforms the baseline WMMSE-Tr (WMMSE truncated to 4 iterations) in sum rate as training proceeds, showing the value of the learnable unfolded architecture. 
The effectiveness of distributed training is demonstrated by its similar sum rate as that of the centralized training. 
 
\subsection{Graph-based Policy Gradient Descent for Link Scheduling}
\label{ssec:link_scheduling}

We train GCNNs applied to distributed link scheduling in wireless multihop networks with orthogonal access~\cite{zhao2023graphbased,zhao2022twc}. 
This task is formulated as finding the maximum weighted independent set (MWIS) on the conflict graph of a wireless network, in which each vertex represents a wireless link (with weights representing the utility of scheduling that link), and an edge indicates that two links cannot be activated at the same time.
The MWIS problem is known to be NP-hard~\cite{Joo2010scheduling_complexity}, and heuristics are used in practical link schedulers.
In this application, an actor GCNN is trained to indirectly improve the quality of the solution by modifying the input vertex weights of a distributed local greedy solver (LGS)~\cite{joo2012local}, which guarantees that the solution is always  an independent set.

We follow the configuration and training process in~\cite{zhao2023graphbased}, which involves alternatively training a 5-layer GCNN (twin) and a 3-layer actor GCNN with 
a set of random graphs drawn from the  Erd\H{o}s-R\'enyi model
with 100 nodes and average node degree ranging from {$2$ to $25$}.
Within each mini-batch ($B=100$), the graph remains the same, and the vertex weights are drawn online from $\mathbb{U}(0,1.0)$, emulating the online training of GCNN in a dynamic network with a topology that changes per mini-batch.
The actor is trained in a fully distributed manner, while the critic GCNN is trained in a centralized manner. 
We use the Adam optimizer with the learning rate $\alpha=5\times 10^{-5}$, and the remaining hyperparameters are as in~\cite{zhao2023graphbased}. 
In Fig.~\ref{fig:scheduling}, the ratio between the total utility achieved by the GCNN-LGS w.r.t. that of the basic LGS as a function of the number of mini-batches is presented. 
Although the GCNN-LGS trained in a fully distributed manner underperforms its centralized counterpart, it still outperforms the baseline LGS, demonstrating the effectiveness of our distributed training approach in a more challenging graph-based ML pipeline. 



\section{Conclusion and Future Steps}\label{sec:conclusion}

We presented a methodology of online training for GNNs applied to fully distributed networked systems.
This approach was illustrated with examples of GCNNs, including local implementations of inference and backpropagation, as well as communication-efficient mini-batch training based on information reuse, distributed gradient descent algorithms, and message piggybacking.
The effectiveness of our approach was demonstrated numerically in supervised, unsupervised, and reinforcement learning for GCNNs in wireless ad-hoc networks. 
Future work includes theoretical proofs of convergence, and studying the impacts of communication errors and network mobility on training performance, and potential improvements.

\bibliographystyle{ieeetr}
{\footnotesize
\bibliography{refs2}

\begin{thebibliography}{10}

\bibitem{wu2020comprehensive}
Z.~Wu, S.~Pan, F.~Chen, G.~Long, C.~Zhang, and S.~Y. Philip, ``A comprehensive survey on graph neural networks,'' {\em IEEE Trans. on Neural Networks and Learning Systems}, 2020.

\bibitem{chien2024opportunities}
E.~Chien, M.~Li, A.~Aportela, K.~Ding, S.~Jia, S.~Maji, Z.~Zhao, J.~Duarte, V.~Fung, C.~Hao, {\em et~al.}, ``Opportunities and challenges of graph neural networks in electrical engineering,'' {\em Nature Reviews Electrical Engineering}, vol.~1, no.~8, pp.~529--546, 2024.

\bibitem{nasir2019marl}
Y.~S. Nasir and D.~Guo, ``Multi-agent deep reinforcement learning for dynamic power allocation in wireless networks,'' {\em IEEE J. Sel. Areas Commun.}, vol.~37, no.~10, pp.~2239--2250, 2019.

\bibitem{chowdhury2021uwmmse}
A.~Chowdhury, G.~Verma, C.~Rao, A.~Swami, and S.~Segarra, ``Unfolding {WMMSE} using graph neural networks for efficient power allocation,'' {\em IEEE Trans. Wireless Commun.}, vol.~20, no.~9, pp.~6004--6017, 2021.

\bibitem{zhao2022twc}
Z.~Zhao, G.~Verma, C.~Rao, A.~Swami, and S.~Segarra, ``Link scheduling using graph neural networks,'' {\em IEEE Trans. Wireless Commun.}, vol.~22, no.~6, pp.~3997--4012, 2023.

\bibitem{zhao2024bbp}
Z.~Zhao, B.~Radojičić, G.~Verma, A.~Swami, and S.~Segarra, ``Biased backpressure routing using link features and graph neural networks,'' {\em IEEE Trans. Mach. Learn. Commun. Netw.}, vol.~2, pp.~1424--1439, 2024.

\bibitem{zhao2024offloading}
Z.~Zhao, J.~Perazzone, G.~Verma, and S.~Segarra, ``Congestion-aware distributed task offloading in wireless multi-hop networks using graph neural networks,'' in {\em IEEE Int. Conf. on Acoustics, Speech and Signal Process. (ICASSP)}, pp.~8951--8955, 2024.

\bibitem{zhao2023graphbased}
Z.~Zhao, A.~Swami, and S.~Segarra, ``Graph-based deterministic policy gradient for repetitive combinatorial optimization problems,'' in {\em Intl. Conf. Learn. Repres. (ICLR)}, 2023.

\bibitem{shen2023gnnforwireless}
Y.~Shen, J.~Zhang, S.~H. Song, and K.~B. Letaief, ``Graph neural networks for wireless communications: From theory to practice,'' {\em IEEE Trans. Wireless Commun.}, vol.~22, no.~5, pp.~3554--3569, 2023.

\bibitem{li2024glance}
B.~Li, G.~Verma, T.~Efimov, A.~Kumar, and S.~Segarra, ``{GLANCE}: Graph-based learnable digital twin for communication networks,'' {\em arXiv preprint arXiv:2408.09040}, 2024.

\bibitem{verbraeken2020dml}
J.~Verbraeken, M.~Wolting, J.~Katzy, J.~Kloppenburg, T.~Verbelen, and J.~S. Rellermeyer, ``A survey on distributed machine learning,'' {\em ACM Comput. Surv.}, vol.~53, Mar. 2020.

\bibitem{mcmahan2017communication}
B.~McMahan, E.~Moore, D.~Ramage, S.~Hampson, and B.~A.~y. Arcas, ``Communication-efficient learning of deep networks from decentralized data,'' in {\em Intl. Conf. Artif. Intel. Stat. (AISTATS)}, vol.~54, pp.~1273--1282, PMLR, Apr 2017.

\bibitem{nedic2009distributed}
A.~Nedic and A.~Ozdaglar, ``Distributed subgradient methods for multi-agent optimization,'' {\em IEEE Trans. Auto. Control}, vol.~54, no.~1, pp.~48--61, 2009.

\bibitem{swenson2022distributed}
B.~Swenson, R.~Murray, H.~V. Poor, and S.~Kar, ``Distributed stochastic gradient descent: Nonconvexity, nonsmoothness, and convergence to local minima,'' {\em J. Mach. Learn. Res.}, vol.~23, no.~328, pp.~1--62, 2022.

\bibitem{nazari2022dadam}
P.~Nazari, D.~A. Tarzanagh, and G.~Michailidis, ``{DADAM}: A consensus-based distributed adaptive gradient method for online optimization,'' {\em IEEE Trans. Signal Process.}, vol.~70, pp.~6065--6079, 2022.

\bibitem{chen2023dams}
X.~Chen, B.~Karimi, W.~Zhao, and P.~Li, ``On the convergence of decentralized adaptive gradient methods,'' in {\em Asian Conf. Mach. Learn. (ACML)}, vol.~189 of {\em Proceedings of Machine Learning Research}, pp.~217--232, PMLR, Dec 2023.

\bibitem{Lin2023survey}
H.~Lin, M.~Yan, X.~Ye, D.~Fan, S.~Pan, W.~Chen, and Y.~Xie, ``A comprehensive survey on distributed training of graph neural networks,'' {\em Proceedings of the IEEE}, vol.~111, no.~12, pp.~1572--1606, 2023.

\bibitem{Shao2024distributed}
Y.~Shao, H.~Li, X.~Gu, H.~Yin, Y.~Li, X.~Miao, W.~Zhang, B.~Cui, and L.~Chen, ``Distributed graph neural network training: A survey,'' {\em ACM Comput. Surv.}, vol.~56, Apr. 2024.

\bibitem{gama2019gcnn}
F.~Gama, A.~G. Marques, G.~Leus, and A.~Ribeiro, ``Convolutional neural network architectures for signals supported on graphs,'' {\em IEEE Trans. Signal Process.}, vol.~67, no.~4, pp.~1034--1049, 2019.

\bibitem{kipf2017semisupervised}
T.~N. Kipf and M.~Welling, ``Semi-supervised classification with graph convolutional networks,'' in {\em Intl. Conf. Learn. Repres. (ICLR)}, 2017.

\bibitem{xiao2006distributed}
L.~Xiao, S.~Boyd, and S.~Lall, ``Distributed average consensus with time-varying metropolis weights,'' {\em Automatica}, vol.~1, pp.~1--4, 2006.

\bibitem{3gppTR38901}
3rd Generation Partnership Project~(3GPP), ``Technical report 38.901: Study on channel model for frequencies from 0.5 to 100 {GHz},'' tech. rep., ETSI, 2020.
\newblock Release 16.

\bibitem{Joo2010scheduling_complexity}
C.~Joo, G.~Sharma, N.~B. Shroff, and R.~R. Mazumdar, ``On the complexity of scheduling in wireless networks,'' {\em EURASIP J.Wireless Commun. and Netw.}, vol.~2010, p.~418934, Sep 2010.

\bibitem{joo2012local}
C.~{Joo} and N.~B. {Shroff}, ``Local greedy approximation for scheduling in multihop wireless networks,'' {\em IEEE Trans. on Mobile Computing}, vol.~11, no.~3, pp.~414--426, 2012.

\end{thebibliography}
}



\end{document}